# Micro-Data Learning: The Other End of the Spectrum

by Jean-Baptiste Mouret (Inria)

*Many fields are now snowed under with an avalanche of data, which raises considerable challenges for computer scientists. Meanwhile, robotics (among other fields) can often only use a few dozen data points because acquiring them involves a process that is expensive or time-consuming. How can an algorithm learn with only a few data points?*

Watching a child learn reveals how well humans can learn: a child may only need a few examples of a concept to "learn it". By contrast, the impressive results achieved with modern machine learning (in particular, by deep learning) are made possible largely by the use of huge datasets. For instance, the ImageNet database used in image recognition contains about 1.2 million labelled examples; DeepMinds's AlphaGo used more than 38 million positions to train their algorithm to play Go; and the same company used more than 38 days of play to train a neural network to play Atari 2600 games, such as Space Invaders or Breakout.

Like children, robots have to face the real world, in which trying something might take seconds, hours, or days. And seeing the consequence of this trial might take much more. When robots share our world, they are expected to learn like humans or animals, that is, in far fewer than a million trials. Robots are not alone to be cursed by the price of data: Any learning process that involves physical tests or precise simulations (e.g., computational fluid dynamics) comes up against the same issue. In short, while data might be abundant in the virtual world, it is often a scarce resource in the physical world. I refer to this challenge as "micro-data" learning (see Figure 1).

The first precept of micro-data learning is to choose as wisely as possible what to test next (active learning). Since computation tends to become cheaper every year, it is often effective to trade data resources for computational resources, that is, to employ computationally intensive algorithms to select the next data point to acquire. Bayesian optimisation [1] is such a data-efficient algorithm that has recently attracted a lot of interest in the machine learning community. Using the data acquired so far, this algorithm creates a probabilistic model of the function that needs to be optimised (e.g., the walking speed of a robot or the lift generated by an airfoil); it then exploits this model to identify the most promising points of the search space. It can, for example, find good values for the gait of a quadruped robot (Sony Aibo / 15 parameters to learn) in just two hours of learning.

The second precept of micro-data learning is to exploit every bit of information from each test. For instance, when a robotic arm tries to reach a point in space, the learning algorithm can perform the movement, then, at the end of the trial, measure the distance to the target. In this case, each test corresponds to a single data point. However, the algorithm can also record the position of the "hand" every 10ms, thus getting thousands of data points from a single test. This is a very effective approach for learning control strategies in robotics; for example, the Pilco algorithm can learn to balance a non-actuated pole on an actuated moving cart in 15-20 seconds (about 3-5 trials) [2].

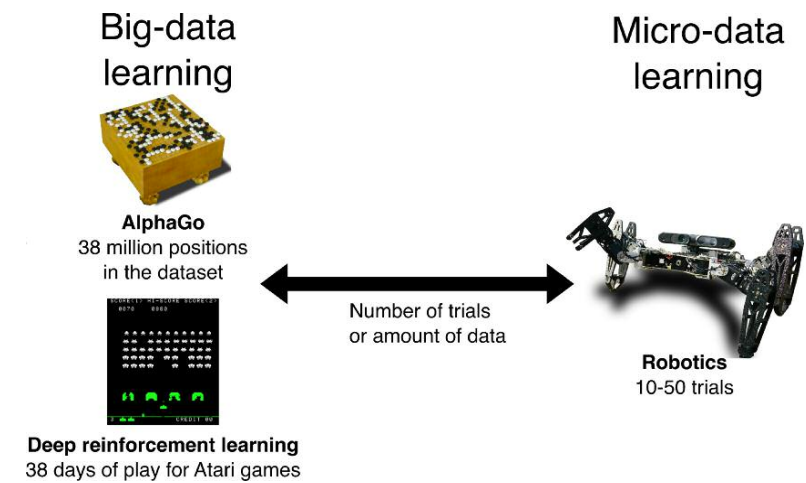

*Figure 1: Modern machine learning (e.g., deep learning) is designed to work with a large amount of data. For example, the Go player AlphaGo by DeepMind used a dataset of 38 million positions, and the deep reinforcement learning experiments from the same team used the equivalent of 38 days to learn to play Atari 2600 video games. Robotics is at the opposite end of the spectrum: most of the time, it is difficult to perform more than a few dozen trials. Learning with such a small amount of data is what we term "Micro-data learning".*

The third precept of micro-data learning is to use the "right" prior knowledge. Most problems are indeed simply too hard to be learned from scratch in a few trials, even with the best algorithms: The quick learning ability of humans and animals is due largely to their prior knowledge about what could and could not work. When using priors, it is critical to make them as explicit as possible, and to make sure that the learning algorithm can question or even ignore them. In academic examples, it can also be challenging to distinguish between prior knowledge that is useful and prior knowledge that actually gives the solu-tion to the algorithm, which leaves nothing to learn.

We focused on prior knowledge in our recent article about damage recovery in robotics [3, L1]. In this scenario, a six-legged walking robot needs to discover a new way to walk by trial-and-error because it is damaged. Before the mission, a novel algorithm explores a large search space with a simulation of the intact robot to identify the most promising solution of each "family". Metaphorically, this algorithm takes the needles out of a haystack to make a stack of needles. If the robot is damaged, the learning algorithm, which is a derivative of Bayesian optimisation [1], exploits this prior knowledge to choose the best trials. In our experiments, the robot discovers compensatory gaits in less than two minutes and a dozen trials, for the five damage conditions that we tested [3].

In this learning approach, a data-efficient learning algorithm that works with the physical, damaged robot is guided by prior knowledge based on a simulation of the intact robot. This micro-data learning algorithm makes it possible to learn a complex task in only a few trials. The subsequent challenge is to exploit more knowledge from the trials [2] and select the next trials while taking the context into account (e.g., potential obstacles).

**Link:**
[L1] http://www.resibots.eu

**References:**
[1] B. Shahriari, et al.: "Taking the human out of the loop: A review of bayesian optimization", Proc. of the IEEE, 2016.
[2] M. P. Deisenroth, D. Fox, C. E. Rasmussen: "Gaussian processes for data-efficient learning in robotics and control", IEEE Trans. on Pattern Analysis and Machine Intelligence, 2016.
[3] A. Cully, et al.: "Robots that can adapt like animals", Nature, 2015.

**Please contact:**
Jean-Baptiste Mouret
Inria, France
jean-baptiste.mouret@inria.fr